\documentclass[runningheads]{llncs}
\usepackage{graphicx}
\usepackage[misc]{ifsym} % marking corresponding author
\usepackage{amsmath}
\usepackage{amssymb}
\usepackage{booktabs}

\usepackage[utf8]{inputenc} % allow utf-8 input
\usepackage[T1]{fontenc}    % use 8-bit T1 fonts
\usepackage{hyperref}       % hyperlinks
\usepackage{url}            % simple URL typesetting
\usepackage{booktabs}       % professional-quality tables
\usepackage{amsfonts}       % blackboard math symbols
\usepackage{nicefrac}       % compact symbols for 1/2, etc.
\usepackage{microtype}      % microtypography
\usepackage{xcolor}         % colors
\usepackage{tikz}
\usepackage{bm}
\usepackage{colortbl}
\usepackage{multirow}
\usepackage{tablefootnote}
\usepackage{wrapfig}
\usepackage{algorithm}
\usepackage{algorithmic}
\usepackage{multicol}
\usepackage[percent]{overpic}
\begin{document}
\title{SGD with Partial Hessian for Deep Neural Networks Optimization}

\author{Ying Sun \inst{1}
\and Hongwei Yong \inst{1}
\and Lei Zhang \inst{1}\Letter}
\authorrunning{Y. Sun et al.}
\institute{The Hong Kong Polytechnic University, Hong Hum, Kowloon, Hong Kong, China. \email{csysun@comp.polyu.edu.hk, hongwei.yong@polyu.edu.hk, cslzhang@comp.polyu.edu.hk}}
%
% %
% %\titlerunning{Abbreviated paper title}
% % If the paper title is too long for the running head, you can set
% % an abbreviated paper title here
\maketitle              % typeset the header of the contribution

\begin{abstract}
    Due to the effectiveness of second-order algorithms in solving classical optimization problems, designing second-order optimizers to train deep neural networks (DNNs) has attracted much research interest in recent years. However, because of the very high dimension of intermediate features in DNNs, it is difficult to directly compute and store the Hessian matrix for network optimization. Most of the previous second-order methods approximate the Hessian information imprecisely, resulting in unstable performance. In this work, we propose a compound optimizer, which is a combination of a second-order optimizer with a precise partial Hessian matrix for updating channel-wise parameters and the first-order stochastic gradient descent (SGD) optimizer for updating the other parameters. We show that the associated Hessian matrices of channel-wise parameters are diagonal and can be extracted  directly and precisely from Hessian-free methods. The proposed method, namely SGD with Partial Hessian (SGD-PH), inherits the advantages of both first-order and second-order optimizers. Compared with first-order optimizers, it adopts a certain amount of information from the Hessian matrix to assist optimization, while compared with the existing second-order optimizers, it keeps the good generalization performance of first-order optimizers. Experiments on image classification tasks demonstrate the effectiveness of our proposed optimizer SGD-PH. The code is publicly available at \url{https://github.com/myingysun/SGDPH}.
\end{abstract}

\section{Introduction}
Owing to the back-propagation algorithm, the first-order local information (e.g., the gradients) of the loss function for deep neural networks (DNNs) can be obtained easily at a reasonable cost, which greatly assists the successful development of the first-order optimizers for training DNNs. 
Among the existing optimizers, the most widely used algorithms are SGD with momentum (SGDM)~\cite{qian1999sgdm} and ADAM~\cite{kingma2014adam}, the descent directions of which are only decided by the gradients.
Because of the advantages including simple implementation, low computational cost and small memory consumption, first-order optimizers~\cite{qian1999sgdm,kingma2014adam,zhuang2020adabelief} occupy the mainstream in the optimization of DNNs.
Live up to expectations, they have been shown the efficiency on a variety of tasks in computer vision~\cite{ren2015faster,he2017mask},  natural language processing~\cite{luong2015effective} and other machine learning areas~(e.g., \cite{mnih2015human,silver2016mastering}).

Besides first-order information, the second-order information (e.g., Hessian) is admittedly considered to help the optimization. Sometimes, second-order algorithms will be preferred in solving traditional optimization problems due to their faster (local) convergence under some assumptions and higher accuracy with fewer iterations. Hence, the generalizations of the second-order methods into DNNs optimization are always under the research spotlight over the past few years.
However, since the dimension of Hessian in DNNs is very high, it is difficult to directly store and compute the Hessian matrix. Therefore, how to make it practical in deep learning is still an open question. A lot of works focus on how to approach the second-order information in DNNs. For example, {some earlier works \cite{martens2010deep,yao2020adahessian,ma2020apollo} apply the Hessian-free methods to approximate the Hessian matrix and then embed this information in the optimization process.}
{Limited by the ways of implementation, most of these methods extract the imprecise second-order information, and the accuracy of these approximations may be considered to affect the performance of the optimizer.}
Instead of the Hessian matrix, the natural gradient matrix can also be considered as the second-order information to be approximated, for example, KFAC~\cite{grosse2016kronecker} and EKFAC~\cite{george2018fast} methods approximate the natural gradient layerwisely by using a block-diagonal of the Fisher matrix. 
{However, these methods are still based on some assumptions about the statistical properties of the parameter distributions, which may also bring inaccuracies into the optimization process.}
Due to such factors, sometimes the performances of the second-order methods are even worse than the first-order ones, which may also hinder the practical application of second-order optimizers.

It is worth noting that the existing optimizers treat all the network parameters ``the same", in other words,  all the parameters in the network follow the same updating rule in the training process. However, we notice that the channel-wise one-dimensional (1D) parameters are commonly introduced in some popular basic modules of DNNs (e.g., in sundry normalization layers). From the perspective of designing optimizers, we find these channel-wise 1D parameters can be proved to have an important property, i.e., the Hessian matrix related to any one group of channel-wised 1D parameters is diagonal (see the derivation in Section \ref{sec3.1}). The dimension of these parameters, which is exactly the number of output channels of this layer, is usually a very small number (e.g., 64), and for these diagonal Hessian matrices, we can obtain the diagonals directly and precisely by the Hessian-free approach (see Section \ref{sec_hf} for details). This conclusion motivates us to treat the parameters ``differently" and propose a new algorithm to combine the first-order methods with the second-order ones.

In this paper, we propose a new type of compound optimizer, named {\bf SGD} with {\bf P}artial {\bf H}essian (SGD-PH), which combines the second-order optimizer for the channel-wise 1D parameters with the first-order optimizer for other parameters.
Through making use of the special structure we mentioned above, we are allowed to use Newton-type methods to update parameters for some specific layers, such as the batch normalization (BN) layer and the convolutional layer with weight normalization (WN).
Compared with first-order optimizers, SGD-PH adopts partial but precise information from the Hessian matrix to help optimization, while compared with other second-order optimizers, it can keep and even surpass the high generalization performance of first-order optimizers in many tasks.
Numerical experiments on different datasets are given to illustrate the effectiveness of our SGD-PH.

%----------------------------------------------------------------------------------------
%------------------------------------------------------------------------------
\section{Related Works}\label{sec1}

When employing second-order methods for DNNs, an essential part is to consider how to extract the Hessian information efficiently and precisely. In this section, we list some Hessian-free methods for solving the descent directions. Besides, we also give a brief introduction of several commonly used normalization methods that inspire our design.

% \noindent
\paragraph{\bf{Hessian-free Approaches:}}
Hessian-free methods are intuitive and efficient in solving the descent direction when applying the Newton-type methods in DNNs optimization. Owing to the advantages of saving storage space, they are super suitable for deep networks and large-scale tensors. One kind of the Hessian-free method approximates the diagonal Hessian elements via back-propagation, e.g., the implementation process in Adahessian \cite{yao2020adahessian} and Apollo \cite{ma2020apollo}. Specifically, Adahessian applies Hutchinson's method as an inexact approximation with the help of Rademacher distribution, and Apollo updates the descent direction by the quasi-Newton method with the help of the weak secant equation, in which the weak secant equation provides a reasonable diagonal approximation of the next Hessian matrix.
Another kind is a generalization of the classical iterative methods, including the Gauss-Seidel method, 
the (preconditioned) conjugate gradient (CG) method, and the generalized minimal residual (GMRES) method. These well-known methods can iteratively solve the linear system without storing the whole Hessian matrix, and have also been embedded into the neural network training pipeline, e.g. \cite{chandra2016fast,martens2010deep}.

% \noindent
\paragraph{\bf{Normalization Methods:}}
In DNNs, there are many specific layers designed to be channel-wise, which enables some good properties for designing a partial Hessian optimizer. The most intuitive examples are the batch normalization (BN)\cite{ioffe2015batch} and weight normalization (WN)\cite{salimans2016weight}.
BN is a usually adopted technique in high-level tasks training. 
In BN layers, a pair of parameters $(\gamma, \beta)$ is introduced to scale and shift  the normalized value by $y_i = \gamma \hat{x}_i + \beta$ for each channel, such that the effect of noise will be reduced. 
By adding BN, the performance of many DNNs become more robust to the change of hyperparameter values.
Differently, WN is a technique that decouples the length and the direction of the weight parameters, i.e., in mathematics, let $\bf{w} = \gamma\frac{\bf{v}}{\|\bf{v}\|}$. By applying the WN method, we are able to decouple the weight length $\gamma$ and the direction tensor $\bf{v}$ from the weight $\bf{w}$. Without calculating statistics from mini-batches like BN, WN is more suitable for some specific applications such as deep reinforcement learning or generative models. Numerous experiments have confirmed that they can both speed up the training process and improve the final generalization performance.
Besides WN and BN, there are also other feature normalization methods such as layer normalization (LN)~\cite{ba2016layer}, group normalization (GN)~\cite{wu2018group} and instance normalization (IN)~\cite{ulyanov2016instance}. Although these normalization layers are simple affine layers, they have a great favorable impact on training deep neural networks, which makes them become popular basic modules in DNNs.

\section{Methodology}\label{sec2}

\begin{figure*}[t]
    \setlength{\abovecaptionskip}{-0pt}
    \begin{center}
        \includegraphics[width=1\linewidth]{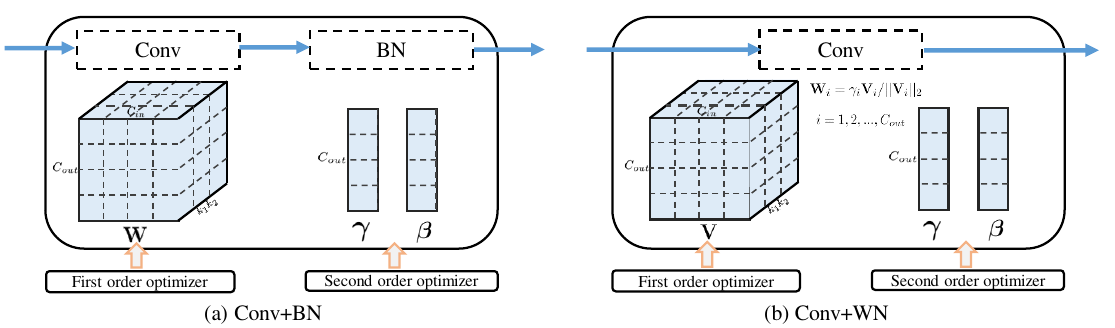}
    \end{center}
    \vspace{-1mm}
    \caption{Illustration of SGD-PH. Here, Figure (a) represents the case when BN is adopted in neural networks training, while other normalization methods such as LN, GN and IN can be represented in the same way. Figure (b) illustrates the case of decoupling the convolutional layers when there are no normalization layers followed, where $\boldsymbol{\beta}$ represents bias, please see Section \ref{sec_general_conv} for more details.}
    \label{fig1}
    \vspace{-2mm}
\end{figure*}

Instead of using the overall first-order or second-order information, we design the descent steps with different order information at different layers, which is the reason we called our method a ``partial Hessian" method. Here, based on the widely applied first-order optimizer SGDM, we add the precise Hessian information when optimizing the 1D parameters.
Moreover, for DNNs that have no normalization layers, we can decouple the convolutional layers into a convolution operation and a linear weight normalization operation, which enables our optimizer to be applied.
Here we use Figure \ref{fig1} to illustrate our idea.
%------------------------------------------------------------
% Reminder: Insert Figure 1 here.

%------------------------------------------------------------------------------------
\subsection{Diagonal Hessian Matrix}\label{sec3.1}

For an intuitive explanation, we adopt the BN layer as an example to describe the design of SGD-PH. For a BN layer, the 1D-parameters are formed by different values of parameters related to different channels.
First we recall the operations of a single BN layer with $C$ input channels. For a batch of input $\mathcal{X} \in \mathbb{R}^{N\!\times\! C \!\times \!W\!\times \!H}$ to this layer, divide it into $C$ sets $\mathcal{X}_1,\ldots, \mathcal{X}_C$ according to the channels, where $\mathcal{X}_i = \{x_i^1, \cdots, x_i^{m_i}\}$ for each channel index $i\in\{1,2,\ldots, C\}$ and $m_i = N W H$ represents the number of the elements within the $i$-th channel.
Then for a specific channel $\mathcal{X}_i$, the BN layer contains the following operations
% \vspace{-1mm}
\begin{equation}
 \mu_{\mathcal{X}_i} = \frac{1}{m_i}\sum_{j=1}^{m_i} x_i^j; \ 
\sigma_{\mathcal{X}_i}^2 = \frac{1}{m_i}\sum_{j=1}^{m_i} (x_i^j- \mu_{\mathcal{X}_i})^2;\ 
 \hat{x}_i^j = \frac{x_i^j - \mu_{\mathcal{X}_i}}{\sqrt{\sigma_{\mathcal{X}_i}^2 + \epsilon}};\ 
y_i^j = {\color{red}\gamma_i} \hat{x}_i^j + {\color{red}\beta_i}.
% \vspace{-1mm}
\end{equation}
The parameters $\gamma_i$, $\beta_i$ introduced here are usually updated via back-propagation in the training process. 
Notice that the BN layer actually normalizes each channel independently, which means 
the parameters $\gamma_i$, $\beta_i$ of $\mathcal{X}_i$ are irrelevant to
$\gamma_j$, $\beta_j$ of $\mathcal{X}_j$ whenever $j\ne i$. 
Thus, denote $\boldsymbol{\Gamma} = (\gamma_1, \cdots, \gamma_C)^T$
as the group of parameters $\gamma_i$ for all the $C$ channels, the Hessian with respect to
$\boldsymbol{\Gamma}$, denoted by $\mathbf{H}_{\boldsymbol{\Gamma}}$ here, is exactly a diagonal matrix. i.e.,
\vspace{-1mm}
\begin{equation}\label{def-diaghessian}
   \mathbf{H}_{\boldsymbol{\Gamma}} := \frac{\partial^2 \mathcal{L}}{\partial \boldsymbol{\Gamma}^2} = {\rm Diag}\left( \frac{\partial^2 \mathcal{L}}{\partial \gamma_1^2}, \cdots,  \frac{\partial^2 \mathcal{L}}{\partial \gamma_C^2}\right)
\vspace{-1mm}
\end{equation}
with $\frac{\partial^2 \mathcal{L}}{\partial \gamma_i^2} = \sum_{j = 1}^{m_i} \frac{\partial^2 \mathcal{L}}{\partial (y_{i}^j)^2}(\hat{x}_{i}^j)^2$ for $i = 1, \ldots, C$. 
Then the inverse $\mathbf{H}_{\boldsymbol{\Gamma}}^{-1}$, if exists,
can be computed directly by the inverse of each diagonal element.
This structure states that getting the precise diagonal elements of the Hessian with respect to $\boldsymbol{\Gamma}$ is equivalent to obtaining the partial Hessian $\mathbf{H}_{\boldsymbol{\Gamma}}$ precisely, which enables us to apply Newton-type methods easily.
The same conclusion also holds for the Hessian of the group of parameters $\boldsymbol{\beta} = (\beta_1, \cdots, \beta_C)^T$. 
Consequently, we can obtain the exact Hessian matrices of such one-dimensional parameters directly instead of using any approximation method.

It is worth mentioning that the property of partial diagonal Hessian does not hold for nature gradient descent methods (that is, the idea of extracting precise partial diagonal Hessian cannot be generalized to optimizers like KFAC), since the Fisher matrices related to 1D parameters are symmetric positive semidefinite matrices $\mathbb{E}[\mathbf{g} \mathbf{g}^\top]$ with $\mathbf{g}$ being the gradient vector, which cannot be proved to be diagonal.

\subsection{The Hessian-free Approach}\label{sec_hf}

As we mentioned in Section \ref{sec1}, the Hessian-free method can be regarded as a necessary way in designing second-order optimizers due to the memory limitations, and some existing optimizers have adopted Hessian-free methods to approximate the whole Hessian diagonal. Under this situation, it seems that there is no need to extract the diagonal of some specific layers.
However, regarding the truth that the whole Hessian is not diagonal, the inexactness brought by approximating the whole diagonal may influence the optimization process, which may lead to the unsatisfactory performance of the second-order optimizers on some tasks.
As a consequence, by extracting the specific elements,
we can avoid the inexactness of the existed diagonal approximation methods, which helps our optimization.

In our optimizer, we adopt the Hessian-free approach with the help of back-propagation. 
To ensure continuity, we still take the BN layer as an example. Under the diagonal Hessian analysis Eq.\eqref{def-diaghessian}, we set $\mathbf{e_{\Gamma}}$ to be the vector that takes elements $1$ related to the 1D vector $\mathbf{\Gamma}$ and takes elements $0$ at all other places. Then with the help of the second time back-propagation, we can get
\vspace{-1mm}
\begin{equation}\label{eq:diagcomp}
   \mathbf{H}\mathbf{e}_{\boldsymbol{\Gamma}}= \frac{\partial (\mathbf{g}^T \mathbf{e}_{\boldsymbol{\Gamma}})}{\partial \boldsymbol{\Gamma}}, 
\vspace{-1mm}
\end{equation}
where $\mathbf{g}$ is the gradient information computed in the first back-propagation process. 
Then the vector ${\rm diag}(\mathbf{H}_{\boldsymbol{\Gamma}})$ is contained in the corresponding positions of the vector $\mathbf{H}\mathbf{e}_{\boldsymbol{\Gamma}}$. By this Hessian-free approach, we can get the corresponding diagonal matrix $\mathbf{H}_{\boldsymbol{\Gamma}}$ precisely without computing and storing the whole Hessian matrix. Hence, our method avoids the inaccuracy brought by the iterative methods or the approximation methods when computing the Hessian information.
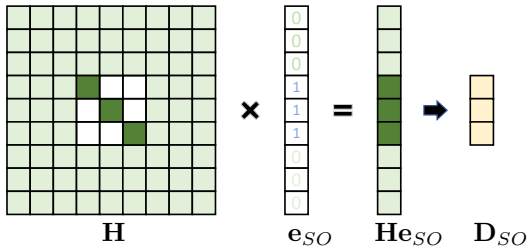
\begin{wrapfigure}{r}{0.6\textwidth}
\begin{center}
    \vspace{-7mm}
    \begin{overpic}[width=0.9\linewidth]{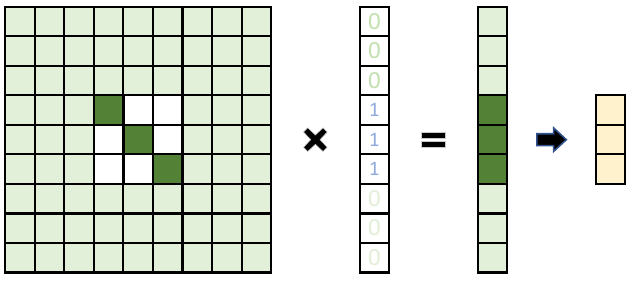}
        \put(20, -4){${\bf{H}}$}
        \put(57.5, -4){${\bf{e}}_{SO}$}
        \put(75, -4){${\bf{H}}{\bf{e}}_{SO}$}
        \put(95, -4){${\bf{D}}_{SO}$}
    \end{overpic}
    \vspace{-1.5mm}
    \end{center}
    \caption{Illustration of diagonal Hessian computation. Here, the light green boxes represent the elements that can be any real numbers while the white boxes represent zeros. The central $3\times 3$ matrix (i.e.,  ${\bf{H}}_{SO}$) in ${\bf{H}}$ is diagonal corresponding to the specific 1D variable. By multiplies with the vector ${\bf{e}}_{SO}$, we can extract the diagonal elements precisely in the middle of ${\bf{H}}{\bf{e}}_{SO}$ and compute the element-wise inverse to get ${\bf{D}}_{SO}$, which is exactly the diagonal of ${\bf{H}}^{-1}_{SO}$.}
    \label{fig2}
\vspace{-4.5mm}
\end{wrapfigure}
The analysis of the group of bias parameters $\boldsymbol{\beta}$ in BN layers is the same as $\mathbf{\Gamma}$. 
Apart from this, other normalization methods, including LN, IN and GN, also have analogous parameters, so similar derivations and conclusions can be obtained.

Generally, for an arbitrary 1D variable in DNN, 
we denote ${\bf{e}}_{SO}$ the vector takes $1$ at positions corresponding to this 1D vector and $0$ at all other positions, and denote the partial diagonal Hessian matrix and the descent direction concerning it by $\mathbf{H}_{SO}$ and $\mathbf{D}_{SO}$, respectively. Figure \ref{fig2} illustrate the idea of our precise diagonal Hessian computation. Through this operation, we can obtain the 
precise partial Hessian for the calculation of descent direction.

\subsection{Techniques in Nonconvex Optimization}

Since a nonconvex objective function may not have a positive-definite Hessian, to apply the Newton method in solving nonconvex stochastic optimization problems, several rectification techniques are usually adopted to guarantee the well-definiteness and the effectiveness of the Newton method.

First, to handle the nonconvexity and noninvertible issues, we use the absolute value matrix with a positive perturbation as the substitution of the partial Hessian matrix, i.e.,
% \vspace{-1mm}
\begin{equation} \label{eq:rectification}
  \tilde{\bf{H}}_{SO} := \sqrt{{\bf{H}}_{SO}^\top {\bf{H}}_{SO}} + \epsilon \bf{I} 
  % \vspace{-1mm}
\end{equation}
with $\epsilon > 0$ a small enough positive real number.
Moreover, at the $t$-th iteration, to minimize the effect of randomness and noise, we apply a
momentum step on the Hessian information $\tilde{\bf{H}}_{SO}$, i.e.,
for some $\alpha \in (0,1)$, the momentum ${\bf{M}}_H$ is computed by 
% \vspace{-1mm}
\begin{equation}\label{eq:momentumhessian}
     {{\bf{M}}_H}^{t} = (1- \alpha){{\bf{M}}_H}^{t-1} + \alpha {\tilde{{\bf{H}}}_{SO}}^{t},
% \vspace{-1mm}
\end{equation}
and the inverse of partial Hessian ${\mathbf{D}_{SO}}^t =({\rm diag}({{\bf{M}}_H}^{t}))^{-1}$ is computed directly by taking the diagonal  element-wise inverse. At each iteration $t$, we only calculate the partial Hessian and the momentum related to the current 1D parameter.
The techniques we mentioned above are some generally used techniques that can also be found in many other papers about second-order optimizers, e.g., \cite{yao2020adahessian,ma2020apollo}.

\noindent
Here, as applied in SGD, we also apply the momentum \cite{qian1999sgdm} calculation on the gradient and the weight decay \cite{krogh1992weightdecay} technique for the final descent direction in our optimizer to accelerate the convergence. Specifically, the momentum step on the gradient is 
% \vspace{-1mm}
\begin{equation} \label{eq:momentumgradient}
   {{\bf{M}}_G}^{t} = (1- \beta){{\bf{M}}_G}^{t-1} + \beta {\bf{G}}^{t},
% \vspace{-1mm}
\end{equation}
where $\beta\in (0,1)$ is a given constant and $\bf{G}$ is the gradient of the parameters. 
Meanwhile, after computing the final descent direction ${\mathbf{D}_G}$ (specifically, taking ${\mathbf{D}_{SO}}{{\bf{M}}_G}$ for 1D parameters and ${\mathbf{M}_G}$ for the others),
for a given weight decay parameter $\eta$ and the weight tensor $\mathbf{W}$, the weight decay step is exactly
  % \vspace{-1mm}
\begin{equation} \label{eq:decoupledwd}
{\tilde{\mathbf{D}}_G}^{t} = {\mathbf{D}_G}^{t} + \eta {\bf{W}}^t.
 % \vspace{-1mm}
\end{equation}
With the techniques in this section, we are ready to introduce the structure of our optimizer SGD with Partial Hessian (SGD-PH) in \textbf{Algorithm 1}.

% --------------------------------------------------------------------------------------------
\begin{algorithm}[t]\label{alg1}
\caption{SGD with partial Hessian (SGD-PH)}
% \LinesNumbered
\hspace*{0.02in} {\bf{Inputs:}} Initial weight vector ${\bf{W}}^0$, step size $\tau$ and $\tau_{SO}$, 
momentum factor $\alpha$ and $\beta$, weight decay parameter $\eta$,
rectification parameter $\epsilon$.\\
\hspace*{0.02in} {\bf{Outputs:}} ${\bf{W}}^{(T)}$.
% ${\bf{W}}^T ={\bf{W}}^T_1, \ldots, {\bf{W}}^T_{\tilde{M}}$.
\begin{algorithmic}[1]
\FOR{$t= 1,\ldots, T$}
\STATE get ${\bf{G}}^t = \nabla L^t({\bf{W}}^t)$;
% \STATE 
\IF{${\bf{W}}^t$ is a channel-wised 1D parameter}
\STATE compute ${{\bf{H}}_{SO}}^t$ and  ${\tilde{\mathbf{H}}_{SO}}^t$ by Eq.\eqref{eq:diagcomp} and Eq.\eqref{eq:rectification}, respectively;
\STATE update ${{\bf{M}}_H}^{t}$ by Eq.\eqref{eq:momentumhessian} and compute diagonal element-wise inverse ${\mathbf{D}_{SO}}^t$; 
\STATE update ${{\bf{M}}_G}^{t}$ by Eq.\eqref{eq:momentumgradient};
\STATE compute ${\mathbf{D}_G}^{t} =\tau_{SO}{\mathbf{D}_{SO}}^t{{\bf{M}}_G}^{t}$;
\ELSE
\STATE update ${{\bf{M}}_G}^{t}$ by Eq.\eqref{eq:momentumgradient};
\STATE give ${\mathbf{D}_G}^{t} = {{\bf{M}}_G}^{t}$;
\ENDIF
\STATE get ${\tilde{\mathbf{D}}_G}^{t}$ by Eq.\eqref{eq:decoupledwd};
\STATE update ${\bf{W}}^{t+1} = {\bf{W}}^t - \tau {\tilde{\mathbf{D}}_G}^{t}$.
\ENDFOR 
\end{algorithmic}
\end{algorithm}

%----------------------------------------------------------------------------
\subsection{Generalizations on Convolutional Layers}\label{sec_general_conv}
In the previous sections, we explained how to apply SGD-PH to DNNs with normalization layers. 
In this section, we will suggest that our optimizer can also be applied in training DNNs without normalization layers in the same way of weight normalization (WN).
Here we illustrate the derivation via a single 2D convolution operation ${\bf{Y}}\! =\! {\bf{W}} \ast {\bf{X}}$, where ${\bf{W}}$ is the kernel with dimension ${C_{out} \times C_{in} \times k_1 \times k_2}$ and ${\bf{X}}$ is the input. 
For each output channel $i\! \in\! \{1, \ldots, C_{out}\}$, the WN operation can be defined as follows
% \vspace{-1mm}
\begin{equation}\label{eq:conv_decouple}
    {\bf{W}}_i = \gamma_i \frac{{{\bf{V}}}_i}{\|{\bf{V}}_i\|_2}, \ \ {\rm where}\ 
    {{\bf{V}}}_i\!\in \!\mathbb{R}^{C_{in}\! \times\! k_1 \!\times\! k_2}\ {\rm and}\ \gamma_i \!\in\! \mathbb{R}.
% \vspace{-1mm}
\end{equation}
Overall, we can get $\mathbf{V} = (\mathbf{V}_1, \ldots, \mathbf{V}_{C_{out}})$ and $\boldsymbol{\Gamma} = (\gamma_1, \ldots, \gamma_{C_{out}})$.
Thus, the parameters to be optimized change from the initial parameter ${\bf{W}}$ to the same dimension parameter  ${\bf{V}}$ and a channel-wised 1D parameter $\boldsymbol{\Gamma}$. Similar to the parameters in BN, our proposed optimization algorithm can also be adopted to optimize the parameter $\boldsymbol{\Gamma}$ with its second-order information. 
Moreover, the bias parameter $\boldsymbol{\beta}$ in the convolutional layers can also be contained into the second-order part optimization of SGD-PH (we omit the details since it is trivial).
We will demonstrate the efficiency of SGD-PH in this generalization case later by the experiments in Section \ref{sec_exp_general_conv}.

\section{Experiments}\label{sec4}

In this section, we will validate the robustness and effectiveness of our proposed SGD-PH by comprehensive experiments on image classification tasks. 
In our experiments, we compare our SGD-PH with first-order optimizers SGDM \cite{qian1999sgdm}, Adam \cite{kingma2014adam}, AdamW \cite{loshchilov2017adamw}, Adabelief  \cite{zhuang2020adabelief}, together with second-order optimizers Adahessian \cite{yao2020adahessian} and Apollo \cite{ma2020apollo}. In Section \ref{ResultsCIFAR}, we accomplish our experiments on deep neural networks VGG11, VGG19 \cite{Simonyan15}, ResNet18 and ResNet50 \cite{he2016resnet} for datasets CIFAR10 and CIFAR100 \cite{krizhevsky2009cifar100},
and in Section \ref{sec:mini}, we compare the performances of different optimizers on ResNet18 and ResNet50 for dataset Mini-ImageNet \cite{vinyals2016miniimagenet}.
To avoid randomness, these experiments are repeated 4 times and the results of testing accuracies are reported in the ``mean $\pm$ std" format. 
Moreover, we report the performance of SGD-PH for the large-scale dataset ImageNet \cite{russakovsky2015imagenet} on ResNet18 in Section \ref{ResultsImageNet}. In Section \ref{sec_exp_general_conv}, we provide an illustration of the generalization of SGD-PH on convolutional layers by the network VGG19.
More ablation studies results of SGD-PH can be found in Section \ref{sec_ablation}.

\paragraph{\bf Experiments Setup:}
In SGD-PH, the hyperparameters $\alpha$ and $\beta$ control the momentum for the partial Hessian matrix and the gradient, respectively. Following the experience from SGDM, we set $\alpha=0.9$ and $\beta=0.9$ in our experiments. The second-order learning rate $\tau_{SO}$ in line 8, Algorithm 1 is set to be $0.001$.
Besides, we add a small positive number $\epsilon=0.0001$ to the Hessian diagonal to avoid the diagonal elements being zero. The batch size, learning rate, weight decay, along with the GPUs we use, will be introduced in each section respectively.

\subsection{Results on CIFAR10/CIFAR100}\label{ResultsCIFAR}

%---------------------------------------------------------------------------------------------------
% Table 4: CIFAR100/CIFAR10
\begin{table*}[t]
% \vspace{-2mm}
\centering
    \caption{Testing accuracies (\%) of different DNNs with different optimizers on CIFAR100/10 datasets.}
    \label{tab:Cifar100}
    \vspace{-2mm}
{\scalebox{0.88}{
 \begin{tabular}{c |c c c c | c c | c}
 \hline %\hline
 \multicolumn{8}{c}{CIFAR100}  \\ \hline 
 {Optimizer} &  {SGDM} & {Adam} & {AdamW} & {Adabelief} & {Adahessian} & {Apollo} & {SGD-PH}  \\ \hline 
 {ResNet18} & $77.20\pm.30$ & $72.95\pm.20$  & $77.23\pm.10$  & $77.43\pm.36$  & $76.73\pm.23$  & {$76.63\pm.27$}   &  $\mathbf{77.96\pm.30}$ \\ 
\rowcolor{gray!25}{ResNet50} & $77.78\pm.43$  & $72.13\pm.53$ & $78.10\pm.17$  & $79.08\pm.23$  & $78.48\pm.22$   &  {$78.68\pm.11$}  & $\mathbf{79.54\pm.27}$ \\ 
 {VGG11} & $70.80\pm.29$  & $68.00\pm.21$ & $71.20\pm.29$  & $72.43\pm.16$  &  $67.78\pm.34$ &  {$70.05\pm.11$}  &  $\mathbf{72.75\pm.13}$ \\ 
\rowcolor{gray!25}{VGG19} & $70.94\pm.32$  & $63.90\pm1.62$ & $70.26\pm.23$  & $72.37\pm.19$  &  $69.93\pm.84$ &  {$71.46\pm.52$}  &  $\mathbf{73.87\pm.28}$ \\ \hline
 \multicolumn{8}{c}{CIFAR10}  \\ \hline 
 {ResNet18} & $95.10\pm.07$ &  $92.95\pm.25$ & $94.80\pm.10$ & $\mathbf{95.12\pm.14}$ & $94.70\pm.15$  & $95.03\pm.12$   &  $95.04\pm.07$ \\ 
\rowcolor{gray!25}{ResNet50} & $94.75\pm.30$ & $92.62\pm.19$  & $94.72\pm.10$ & $\mathbf{95.35\pm.05}$ & $\mathbf{95.35\pm.11}$  &  $95.27\pm.11$  & $95.22\pm .07 $  \\ 
 {VGG11} & $92.17\pm.19 $ &  $90.75\pm.15$ & $92.02\pm.08 $ & $92.45\pm.18$ & $91.85\pm.16$  &  $92.38\pm.19$  & $\mathbf{92.60\pm.08}$ \\
\rowcolor{gray!25}{VGG19} & $93.57\pm.13$  & $92.19\pm.07$ & ${93.54\pm.28}$  & $93.72\pm.08$  &  $93.68\pm.14$ &  {$\mathbf{93.76\pm.07}$}  &  ${93.75\pm.10}$ \\ \hline %\hline
\end{tabular}
}}
\vspace{-1mm}
\end{table*}

%%%%%%%%%%-----------------------------------------------------
% Figure 3: CIFAR100
\begin{figure}[t]
    \begin{center}
    \includegraphics[width=1\linewidth, height= 0.47\linewidth, trim=90 0 90 0,clip]{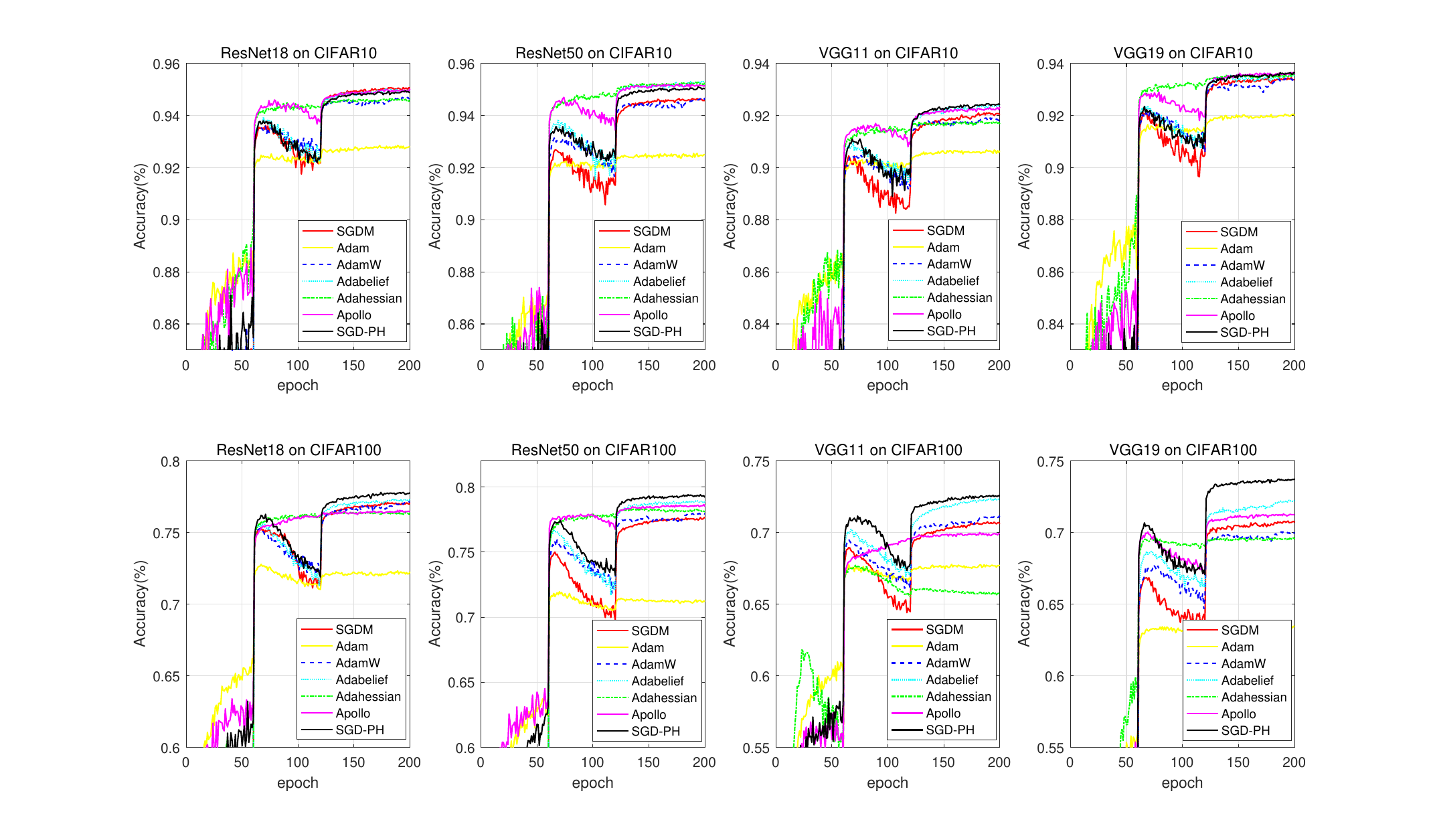}
    \end{center}
      \vspace{-8mm}
    \caption{Testing accuracy curves of different optimizers for different DNNs on CIFAR100 and CIFAR10 datasets.}
    \label{fig_cifar} 
    \vspace{-1mm}
\end{figure}

CIFAR10 and CIFAR100 are commonly used image classification datasets, in which CIFAR10 contains 10 classes of color images for classification with $6000$ images per class, and CIFAR100 contains 100 classes images with $600$ images per class. 
Our experiments here are accomplished on Pytorch 1.7 framework with mainly TITAN RTX and Geforce RTX 2080Ti GPUs. In our experiments, we train DNNs with different optimizers for total $200$ epochs, using batch size $128$ with one single GPU, and the learning rate is multiplied by $0.1$ every $60$ epochs.

\paragraph{\bf Compared with Other Optimizers:}
We compare SGD-PH with four popular first-order optimizers SGDM, Adam, AdamW and Adabelief, and two second-order optimizers Adahessian and Apollo, on four representative DNNs, i.e., ResNet18, ResNet50, VGG11 and VGG19. We tune the learning rate and weight decay for all these methods, and choose their best results for comparison. Specifically, the learning rates are set to be 0.1  for SGDM,  0.001 for Adam, AdamW and Adabelief, $0.15$ for Adahessian and $1$ for Apollo. the  weight decays are set to be $0.0005$ for SGDM, Adam and Adahessian, $0.5$ for AdamW and Adabelief, and $0.00025$ for Apollo.  
% The rest hyperparameters are set the same as their official default settings. 
For SGD-PH, the learning rate takes 0.01 and the weight decay takes 0.005.

Table \ref{tab:Cifar100} shows the testing accuracies with these optimizers on CIFAR100 and CIFAR10.
It can be seen from the results that SGD-PH achieves the best results of all DNN models on CIFAR100. Specifically, SGD-PH surpasses other compared methods largely. The performance of SGD-PH surpasses SGD  from $0.76\%\sim 2.93\%$ on CIFAR100, which fully indicates that the precise partial Hessian can help improve the final performance. For CIFAR10, we notice that many optimizers(e.g., SGDM, Adabelief, Apollo and SGD-PH) reach 100\% training accuracy with loss near to zero, so most of them show quite similar generalization performance as in Table \ref{tab:Cifar100} and the final stage in Figure \ref{fig_cifar}.
In the meantime, we also find that the performance of Adahessian sometimes is not stable, e.g., its performance of VGG on CIFAR100 drops largely. The imprecise Hessian adopted in Adahessian sometimes may have very bad impacts on the optimization process.

Meanwhile, for more intuitive insight, we present the testing accuracy curves of different optimizers on CIFAR100/10 during training different networks in Figure \ref{fig_cifar}. From the trajectories, we see that the performance of Adahessian seems not to gain too much from the learning rate decay at the 120-th epoch. Moreover, affected by the proportion of 1D variables, the trend of the training curves of SGD-PH are usually closer to the first-order optimizers. However, due to the absorption of second-order information, the performance of SGD-PH exists almost always above that of SGDM after the first learning rate decay at the 60-th epoch. This helps to illustrate that SGD-PH may aggregate the advantages of first-order and second-order methods.

\paragraph{\bf Time and Memory Cost:}
{To extract the precise partial Hessian information without approximation, the Hessian free method via the second time back-propagation is the best approach for SGD-PH. In Adahessian, the same technique has also been applied, and sadly the time consumption and the storage cost increase exaggeratedly (e.g., for ResNet18 on CIFAR100, Adahessian has $4.64$ times increase in time and $1.30$ times increase in storage compared to SGDM), which may result in a bad impact on its comprehensive applications. By noticing this fact, we have optimized the calculation process in our Hessian-free approach due to our special structure to shorten the time and storage cost of the second time back-propagation. Although we still have increments of $2.22$ and $1.23$ times in time and storage respectively compared to SGDM due to the properties of back-propagation, we have saved the time and storage a lot compared to Adahessian that adopting the same technique.
}

\subsection{Results on Mini-ImageNet}\label{sec:mini}

\begin{figure}[t]
\begin{minipage}[t]{0.5\textwidth}
    \begin{center}
        \includegraphics[width=\linewidth, trim=59 0 62 0,clip]{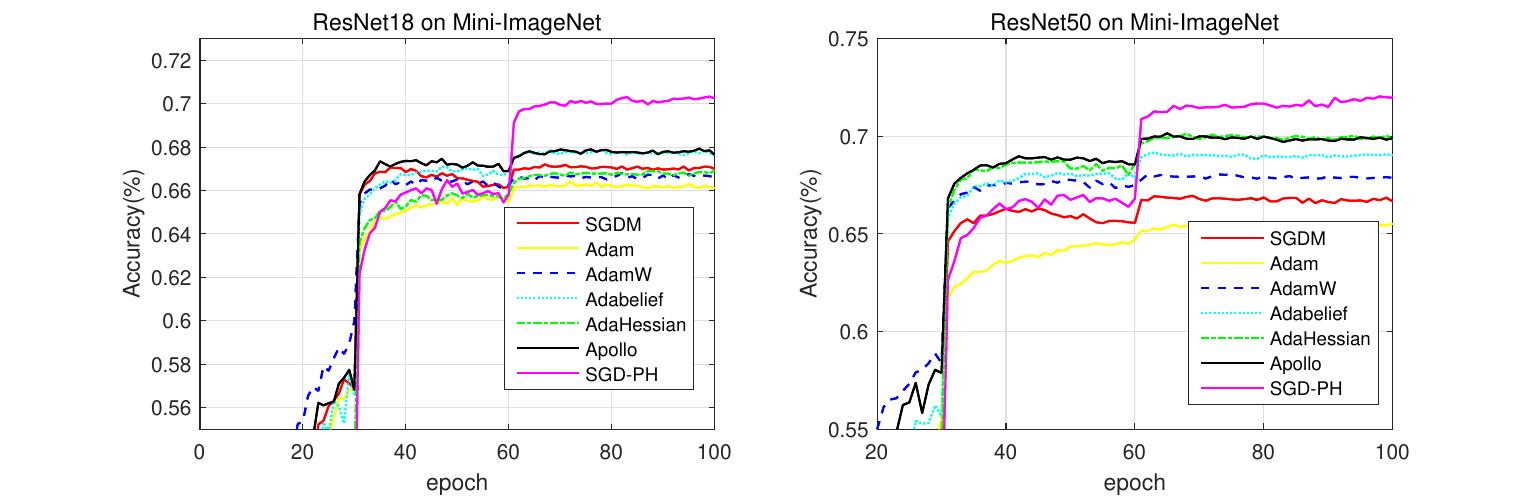}
    \end{center}
    \vspace{-5.5mm}
    \caption{Testing accuracy curves of different optimizers  on Mini-ImageNet dataset.}
    \label{fig:miniimagenet}
\end{minipage}
\hspace{1mm}
\begin{minipage}[t]{0.5\textwidth}
    \begin{center}
        \includegraphics[width=\linewidth, trim=55 0 55 0,clip]{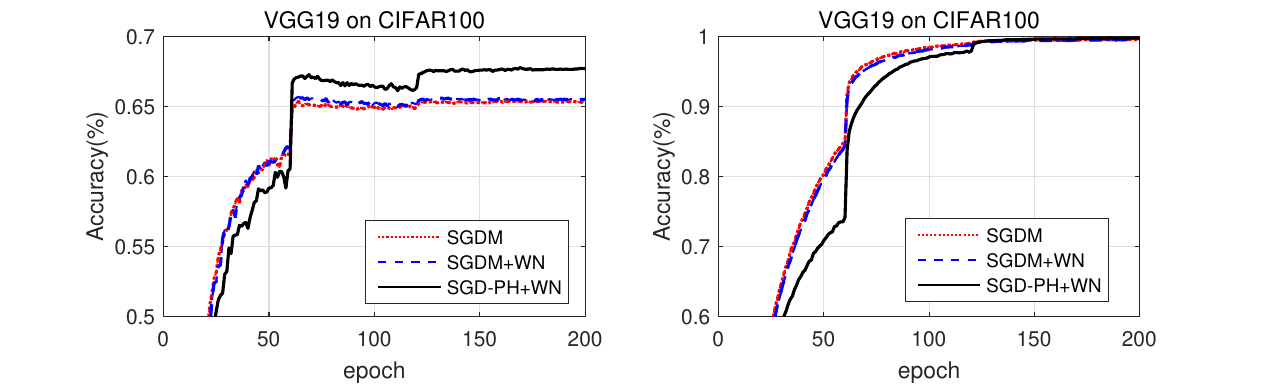}
    \end{center}
    \vspace{-5.5mm}
    \caption{Testing and training accuracy curves of VGG19 without BN layers.}
    \label{fig:WN}
\end{minipage}
\vspace{-3mm}
\end{figure}

Mini-ImageNet is a subset of the well-known dataset ImageNet \cite{russakovsky2015imagenet}. In our tests, 
we use the train/test sets splits provided by \cite{ravi2016optimization,iscen2019label,yong2020gradient}. 
Mini-ImageNet consists of 100 classes and each class has 500 images for training and 100 images for testing. The image resolution is $84\times 84$, and here we resize the images into $224\times 224$ (i.e., the standard ImageNet training input size).  
We train the ResNet18 and ResNet50 with different optimizers for total $100$ epochs with batch size $128$ on  4 GPUs. 
Same as CIFAR100 and CIFAR10 datasets, we repeat each experiment for 4 times to avoid randomness, and the results are reported in the mean $\pm$ std format. The learning rate is multiplied by 0.1 every 30 epochs. For SGD-PH, we set the learning rate to be $0.05$ and the weight decay to be $0.001$. For other optimizers, the learning rates are the same as the settings in CIFAR100/10. The weight decay parameter is set to be $0.0001$ for SGDM, Adam, and Apollo, and $0.1$ for AdamW and Adabelief, $0.0005$ for Adahessian. 
Table \ref{tab:miniimagenet} shows the testing accuracies with these optimizers on Mini-ImageNet, and Fig. \ref{fig:miniimagenet} presents their testing accuracy curves during training. We can see that SGD-PH outperforms other compared methods by a large margin, i.e., $2.47\%$ and  $1.92\%$ performance gains on ResNet18 and ResNet50, respectively. Meanwhile, on mini-ImageNet dataset, the second-order optimizers usually perform better than the first-order optimizers.
Both Adahessian and Apollo achieve favorable generalization performance. Under such cases, SGD-PH can keep the advantages of second-order optimizers, and even further improve their performance. 

% The rest hyperparameters are set identical to their official settings.

% \begin{wrapfigure}{r}{0.6\textwidth}
% \vspace{-2.5mm}
%     \begin{center}
%         \includegraphics[width=\linewidth, trim=59 0 62 0,clip]{mini.eps}
%     \end{center}
%     \vspace{-4mm}
%     \caption{Testing accuracy curves of different optimizers  on Mini-ImageNet dataset.}
%     \label{fig:miniimagenet}
%     \vspace{-4mm}
% \end{wrapfigure}
%----------------------------------------------------------------------------
%%%%%%%%%%%%%%%%%%% table of mini imagenet.
\begin{table*}[t]
\centering
\caption{Testing accuracies (\%) of DNNs with different optimizers for ResNet18/50 on Mini-ImageNet dataset.}
\label{tab:miniimagenet}
\vspace{-1mm}
{\scalebox{0.88}{
 \begin{tabular}{c |c c c c |c c| c}
 \hline %\hline
 % \multicolumn{8}{c}{Mini-ImageNet}  \\ \hline 
 {Optimizer} &  {SGDM} & {Adam} & {AdamW} & {Adabelief} & {Adahessian} &{Apollo} & {SGD-PH}  \\ \hline 
 {ResNet18} & {$67.33 \pm .17$ } &  {$66.47 \pm .34$ } &  { $66.90 \pm .36$ }  &   {$67.98 \pm .29$} & {$67.13 \pm .40$}& {$68.06 \pm .38$}  &  $\mathbf{70.53 \pm .32}$  \\ 
\rowcolor{gray!25}{ResNet50} & {$67.09 \pm .56$ } &  {$65.70 \pm .60$ } &  { $68.23 \pm .30$ }  &   {$69.36 \pm .13$}  &{$70.25 \pm .28$} & {$70.24 \pm .28$}  &   $\mathbf{72.17 \pm .30}$  \\ \hline % \hline
\end{tabular}
}}
\vspace{1.5mm}
\end{table*}

% Table 5: ImageNet.
\begin{table*}[t]
\centering
\caption{Testing accuracies (\%) of DNNs with different optimizers on ImageNet. The result of ADAM, AdamW, Adabelief and Adahessian are cited from \cite{liu2019radam}, \cite{chen2018closing}, \cite{zhuang2020adabelief}, and  \cite{yao2020adahessian}, respectively.}
    \label{tab:results_imagenet}
    \vspace{-2mm}
{\scalebox{0.96}{
 \begin{tabular}{c |c c c c |c c| c}
 \hline 
 % \multicolumn{8}{c}{ResNet18 on ImageNet}  \\ \hline 
 {\ Optimizer\ } &  {\ SGDM\ } & {\ Adam\ } & {\ AdamW\ } & {\ Adabelief\ } & {\ Adahessian\ } &{\ Apollo\ } & {\ SGD-PH\ }  \\ \hline 
 {Accuracy} & {70.49} &  {66.54} &  {67.93}  &   {70.08} & {70.08}& {70.39}  &  $\mathbf{70.59}$ \\ \hline 
\end{tabular}
}}
\vspace{-2.5mm}
\end{table*}

\subsection{Results on ImageNet} \label{ResultsImageNet}
In this section, we report the results on ImageNet \cite{russakovsky2015imagenet} to validate the effectiveness of SGD-PH. 
ImageNet is a large image classification dataset that contains 1000 categories with 1.28 million images for training and 50K images for validation. 
{Our experiments on ImageNet are accomplished on Pytorch 1.7 framework with four GeForce RTX 2080Ti GPUs.} 
In our experiments, we train the network ResNet18~\cite{he2016resnet} for total $100$ epochs with batch size $256$ on 4 GPUs, and the learning rate is multiplied by 0.1 every 30 epochs. 
We test the performance of SGD-PH, SGDM and Apollo, while we cite the performance of other optimizers from existing papers in Table \ref{tab:results_imagenet} for a more intuitive and complete comparison. {We set the initial learning rate to be $0.1$ for SGD-PH and SGDM, and 1 for Apollo, and set the weight decay to be $0.0001$ for all SGD-PH, SGDM and Apollo.} 
The rest hyperparameters all follow their official settings.

Table \ref{tab:results_imagenet} shows the testing accuracy of these optimizers on ImageNet.
Among the experiments on ImageNet reported in the existing papers, the first-order optimizer SGDM usually has a favorable performance, and the generalization ability is often not inferior to other compared optimizers, which means these optimizers sometimes may not perform stably on the large scale dataset.
Meanwhile, our SGD-PH, under the same initial learning rate and weight decay with SGDM, can maintain this stability of performance by $0.1\%$ performance gain compared with SGDM. This is also a reflection of the fact that our SGD-PH can inherit the advantages of the first-order optimizers, which states the capability and universality of the newly proposed SGD-PH for training DNNs on large-scale datasets.

% \begin{figure}[t]
% \begin{minipage}[t]{0.5\textwidth}
%     \setlength{\abovecaptionskip}{-0pt}
%     \begin{center}
%     \includegraphics[width=\linewidth,trim=59 0 62 0,clip]{mini.eps}
%     \end{center}
%     \vspace{-2.5mm}
%     \caption{Testing accuracy curves of different optimizers  on Mini-ImageNet dataset.}
%     \label{fig:miniimagenet}
% \end{minipage}
% \hspace{1mm}
% \begin{minipage}[t]{0.5\textwidth}
%     \setlength{\abovecaptionskip}{-0pt}
%     \begin{center}
%     \includegraphics[width=1\linewidth,trim=55 0 55 0,clip]{WN.eps}
%     \end{center}
%     \vspace{-2.5mm}
%     \caption{Testing and training accuracy curves of VGG19 without BN layers on CIFAR100.}
%     \label{fig:WN}
% \end{minipage}
% \vspace{-3mm}
% \end{figure}

\begin{table}[t]\scriptsize
\centering
\caption{Testing accuracies (\%) of VGG19 with/without normalization layers on CIFAR100 dataset.}
\label{tab:ConvWn}
\vspace{-1mm}
{\resizebox{\textwidth}{!}{
 \begin{tabular}{c |c | c| c| c | c}
 \hline 
 % \multicolumn{6}{c}{VGG19 on CIFAR100}  \\ \hline 
  {\ Optimizer\ } & {\ SGDM\ } & {\ SGDM+WN\ } & {\ SGD-PH+WN\ } & {\ SGDM+BN\ } & {\ SGD-PH+BN\ } \\
 \hline
 {Accuracy} &  { $65.56 \pm .41$ } &  {$65.98 \pm .35$}  & ${67.93 \pm .35}$ 
 &  $70.94\pm.32$ & $\mathbf{73.87\pm.28}$ \\ \hline 
\end{tabular}
}}
\vspace{1mm}
\end{table}

% \begin{wrapfigure}{r}{0.6\textwidth}
% \vspace{-2.5mm}
%     \begin{center}
%         \includegraphics[width=\linewidth, trim=59 0 62 0,clip]{mini.eps}
%     \end{center}
%     \vspace{-4mm}
%     \caption{Testing accuracy curves of different optimizers  on Mini-ImageNet dataset.}
%     \label{fig:miniimagenet}
%     \vspace{-4mm}
% \end{wrapfigure}
% \begin{wrapfigure}{r}{0.6\textwidth}
% \vspace{-8mm}
%     \begin{center}
%         \includegraphics[width=\linewidth, trim=55 0 55 0,clip]{WN.eps}
%     \end{center}
%     \vspace{-6mm}
%     \caption{Testing and training accuracy curves of VGG19 without BN layers.}
%     \label{fig:WN}
%     \vspace{-4.5mm}
% \end{wrapfigure}
%----------------------------------------------------------------------------
\subsection{Results about Convolutional Layers}
\label{sec_exp_general_conv}
For DNNs without normalization layers, we can apply SGD-PH by adopting WN to embed the 1D parameters. As we introduced in Figure \ref{fig2} (b), we reformulate the convolution operation as the composition of a linear operator (corresponding to the length) with a convolution operator (corresponding to the direction) as is often applied in WN. Here we use VGG19 \cite{Simonyan15} as an example to illustrate the generalization of our optimizer on general DNNs. 
In these experiments, all the BN layers of the VGG19 model have been removed. It is well-known that the performance of DNNs may drop largely without BN. We compare SGD and SGD+WN with SGD-PH+WN. We tune both the learning rate and weight decay for these three methods and report their best results. The learning rate is tuned to be 0.01 for them all. The weight decays are 0.0001 for SGD and SGD+WN, and 0.001 for SGD-PH+WN. The other training strategies are the same as those introduced in Section \ref{ResultsCIFAR}. As a more intuitive comparison, we also report the results of the VGG19 model with BN layers in Table \ref{tab:ConvWn} using the same results reported in Section \ref{ResultsCIFAR}.

Table \ref{tab:ConvWn} gives the results of these five methods and  Figure \ref{fig:WN} shows the testing and training accuracy curves during training for the model VGG19 without BN layers.  
It is easy to see that the performance drop largely without BN, e.g., from $70.94\%$ to $65.56\%$ for SGDM. In this case, WN can slightly improve the performance of SGDM, while SGD-PH+WN can largely boost the performance over SGDM+WN by $2.37\%$. 
These results fully demonstrate the effectiveness of our proposed SGD-PH, that is, whether for a DNN with or without BN layers, SGD-PH can gain the final performance.

\subsection{Ablation Studies}\label{sec_ablation}
%------------------------------------------------------------------------------------
% Table 1: lr and wd tuning.
\begin{table}[t]\scriptsize
\centering
\caption{Testing accuracies  (\%) of different LR and WD  for ResNet18 on CIFAR100.}
\label{tab:lr_wd}
\vspace{-2mm}
{\resizebox{0.9\textwidth}{!}{
\begin{tabular}{c |c c c c c}
 \hline 
 % \multicolumn{6}{c}{ResNet18 on CIFAR100}  \\ \hline 
 {LR} &  {0.01} & {0.01} & {0.01} & {0.02} & {0.005}\\ \hline 
 \rowcolor{gray!25} {WD} & 0.01 & 0.002  & 0.005 & 0.005 &  0.005  \\ \hline
 {Accuracy} &\ $77.19\pm.45$\ &\ $77.33\pm.14$\  &\  $\mathbf{77.96\pm.30}$\  &\  $77.21\pm.20$\ &\  $77.28\pm.28$\ \\ \hline 
\end{tabular}
}}
\vspace{1mm}
\end{table}

% %----------------------------------------------------------------------------
% Table 3: \tau_{SO} tuning.
\begin{table}[t]\scriptsize
\centering
\caption{Testing accuracies (\%) of different second-order learning rate $\tau_{SO}$ for different DNNs on CIFAR100/10 datasets.}
\label{tab:tauso}
\vspace{-2mm}
{\resizebox{0.99\textwidth}{!}{
 \begin{tabular}{c |c c c| c c c}
 \hline 
 {Dataset} & \multicolumn{3}{c|}{CIFAR100} & \multicolumn{3}{c}{CIFAR10} \\ \hline 
 {$\tau_{SO}$} &  0.01 &  0.001 & {0.0001} & 0.01 & 0.001 & {0.0001} \\ \hline 
 {ResNet18} & {$77.33 \pm .24$} &  ${77.96\pm.30}$ & $\mathbf{78.22\pm .20}$  & {$\mathbf{94.99\pm .08}$} &  ${94.98\pm.07}$   & $94.86\pm.06$ \\ 
\rowcolor{gray!25}{ResNet50}  & $\mathbf{79.61 \pm .20}$ &  ${79.54\pm.27}$  &  ${79.53\pm.20}$  & $\mathbf{95.27\pm .10}$ &  ${95.17\pm .10} $ &  $95.07\pm .12 $\\ 
 {VGG11}  & {$70.54 \pm .23$} &  ${72.75\pm.13}$ &  $\mathbf{73.21\pm.09}$  & {$92.14 \pm .13$} &  $\mathbf{92.64\pm.24}$ & ${92.58\pm.20}$ \\ 
\rowcolor{gray!25} {VGG19}  & {$72.78\pm.17$} &  $\mathbf{73.87\pm.28}$ &  ${73.62\pm.21}$  & {$93.54\pm.18$} &  $\mathbf{93.77\pm.19}$ & ${93.55\pm.20}$ \\ \hline
\end{tabular}
}}
\vspace{-2mm}
\end{table}

\begin{figure}[t]
\begin{minipage}[t]{0.4\textwidth}
\centering
\makeatletter\def\@captype{table}\makeatother\caption{Testing accuracies (\%) of $\alpha$ for ResNet18 on CIFAR100.}
\vspace{2.5mm}
 \resizebox{\textwidth}{!}{
 \begin{tabular}{c |c  c  c }
 \hline %\hline
  % \multicolumn{4}{c}{ResNet18 on CIFAR100}  \\ \hline 
 {\ $\alpha$\ } &  {0.8} & {0.9} & {0.99} \\ \hline 
{Accuracy} & \ ${77.79 \pm .35}$ \ &\  $\mathbf{77.96\pm .30}$ \ & \ $77.81 \pm .32$ \   \\ \hline %\hline
\end{tabular}\label{tab:somom}}
\end{minipage}
\hspace{0.5mm} 
\begin{minipage}[t]{0.6\textwidth}
\centering
\makeatletter\def\@captype{table}\makeatother\caption{Testing accuracies (\%) of different batch size settings for ResNet18 on CIFAR100.}
\vspace{2.5mm}
 \resizebox{\textwidth}{!}{        
 \begin{tabular}{c |c c c c c }
 \hline 
  % \multicolumn{6}{c}{ResNet18 on CIFAR100}  \\ \hline 
 {Batch Size} &  {16} & {32} & {64} & {128}& {256}  \\ \hline 
 {Accuracy} & \ $78.08\pm 0.08$\ & \ $\mathbf{78.32 \pm .21}$\   & \ $78.14 \pm .35$ \ & \ $77.96\pm .30$ \ & \ $77.56\pm .16$ \ \\ \hline
\end{tabular}\label{tab:bs}}
\end{minipage}
\vspace{-3mm}
\end{figure}

In this section, we will report some ablation studies about SGD-PH to state its robustness and efficiency. 
We first tune some important hyperparameters, including the initial learning rate, weight decay, the second-order learning rate and the momentum of partial Hessian, then we test the performance of SGD-PH under different input batch size settings. Similarly, we repeat each experiment 4 times and report the results in the mean $\pm$ std format.

% \noindent
\paragraph{\bf Learning Rate and Weight Decay:}
Generally, good settings for learning rate and weight decay can greatly benefit the final generalization results.
In our experiments, we choose the learning rate (LR) and weight decay (WD) from the set $\{0.005, 0.01, 0.02\}$, 
while the learning rate of the second-order part $\tau_{SO}$ and the Hessian momentum $\alpha$ are set to be $0.001$ and $0.9$, respectively. Table \ref{tab:lr_wd} shows the results of five combinations of LR and WD. It can be found that the best result is setting LR to be $0.01$ and WD to be $0.005$, and we adopt this in the experiments of Section \ref{ResultsCIFAR}. 
Moreover, other settings also achieve acceptable results with fluctuations of no more than 0.77\%, which illustrates that SGD-PH is robust to these hyperparameters.

% \noindent
\paragraph{\bf Second-order Learning Rate:}
In SGD-PH, an additional important hyperparameter is imported, i.e., the learning rate of the second-order part $\tau_{SO}$. 
Here we tune $\tau_{SO}$ on CIFAR100/10 by traversing through the set $\{0.01, 0.001, 0.0001\}$, while the initial LR and WD are set to be 0.01 and 0.005, respectively.
Table \ref{tab:tauso} gives the testing accuracies of different $\tau_{SO}$ on CIFAR100/10.
We can see from the results that, on CIFAR100, a smaller $\tau_{SO}$ (i.e., $0.0001$) may achieve better generalization performances in some shallow neural networks, e.g., VGG11 and ResNet18. However, for the deeper networks, a small $\tau_{SO}$ may no longer have such an advantage, on the contrary, a larger $\tau_{SO}$ like $0.01$ may perform better. 
Meanwhile, on CIFAR10, the performances of SGD-PH with different $\tau_{SO}$  have no significant differences. Thus, the results in Table \ref{tab:tauso} shows the applicability of our proposed 
SGD-PH related to different second-order LR $\tau_{SO}$. 
Furthermore, as a moderate choice, we take the hyperparameter $\tau_{SO} = 0.001$ in the above sections, and it works well through all of our experiments.

% \noindent
\paragraph{\bf Hessian Momentum:}
Following the experience of tuning the momentum of SGDM, we list some testing accuracy results of SGD-PH with different Hessian momentum parameters $\{0.8, 0.9, 0.99\}$ for ResNet18 on CIFAR100 in Table \ref{tab:somom}.  As shown in Table \ref{tab:somom}, $\alpha = 0.9$ attains the best results (which is the value we adopt in the previous sections), while other parameters also achieve acceptable results with a maximum fluctuation of 0.17\%.
Consequently, SGD-PH performs stably with these commonly chosen values of momentum.

% \noindent
\paragraph{\bf Batch Size:}
The input batch size can also affect the performance of optimizers.
Hence, we also pay attention to the impact of different batch size on the performance of SGD-PH.
Here, we provide Table \ref{tab:bs} about the testing accuracy results of SGD-PH for ResNet18 on CIFAR100, with the hyperparameters settled by lr$=\!0.01$, wd$=\!0.005$, $\tau_{SO} = 0.001$ and $\alpha = 0.9$. When the batch size increases from 16 to 256, there is a certain range of fluctuations in the testing accuracies, with the best result occurring at batch size $32$. Totally, the trend of fluctuation about SGD-PH related to batch size is similar to other widely used optimizers, which ensures the adaptability and stability of SGD-PH for different settings in applications.

%------------------------------------------------------ 
\section{Conclusions}\label{sec_conclusion}
In this paper, we propose SGD-PH, a compound optimizer that combines first-order optimizer SGDM with partially accurate Hessian information. The design of SGD-PH is based on the derivation of the Hessian matrices of the channel-wise 1D parameters, which are proved to be diagonal matrices and can be extracted precisely through the Hessian-free method.
Besides showing the effectiveness of SGD-PH on DNNs with the widely used normalization layers, we also give an example applying it to the reformulated convolutional layer directly, which illustrates that our optimizer can be applied to any DNNs (even if without normalization layers) with a satisfactory performance achieved.
Sufficient ablation studies are accomplished to verify the robustness and adaptability of our proposed SGD-PH related to different hyperparameters.
However, consistent with other second-order optimizers, our SGD-PH also needs more computational time and memory compared with first-order optimizers. This is still one of the key problems faced for designing second-order optimizers.

% \paragraph{\bf Ethics Statements}: As a basic research work that proposes a new optimizer for training DNNs, our contribution lies in the optimizer SGD-PH itself, which does not relate to collecting data and inferring personal information. Currently, this work does not involve any ethical implications.

\bibliographystyle{splncs04}
\bibliography{main.bib}

\end{document}